\documentclass[11pt]{article}

\usepackage{styles/acl}

\usepackage{times}
\usepackage{latexsym}
\usepackage{booktabs}
\usepackage{amstext}
\usepackage{cleveref}
\usepackage{multicol}
\usepackage{multirow}
\usepackage{soul}
\usepackage{lipsum}
\usepackage{diagbox}
\usepackage[usestackEOL]{stackengine}

\usepackage[nolist]{acronym}
\usepackage[T1]{fontenc}

\usepackage[utf8]{inputenc}

\usepackage{microtype}
\usepackage{rotating}
\usepackage{nameref}
\usepackage{wrapfig}
\usepackage{paralist}
\usepackage{makecell}

\title{Resolving Legalese: A Multilingual Exploration of Negation Scope Resolution in Legal Documents}

\author{Ramona Christen$^{1\;}\thanks{\hspace{2mm}Equal contribution.}$  \And
  Anastassia Shaitarova$^{2}$  \And
  Matthias Stürmer$^{1,3}$ \AND
  Joel Niklaus$^{1,3,4\;*}$ 
\\\\
$^1$University of Bern\quad
$^2$University of Zurich\\
$^3$Bern University of Applied Sciences\quad
$^4$Stanford University
}

\begin{document}

\maketitle

\begin{acronym}
\acro{bert}[BERT]{Bidirectional Encoder Representation from Transformers}
\acro{mlm}[MLM]{Masked Language Modeling}
\acro{nsp}[NSP]{Next Sentence Prediction}
\acro{lm} [LM] {language model}
\acrodefplural{lm} [LMs] {language models}
\acro{sota} [SotA] {state-of-the-art}
\acro{fscs}[FSCS]{Federal Supreme Court of Switzerland}
\end{acronym}

\begin{abstract}
Resolving the scope of a negation within a sentence is a challenging NLP task.\ The complexity of legal texts and the lack of annotated in-domain negation corpora pose challenges for \ac{sota} models when performing negation scope resolution on multilingual legal data. Our experiments demonstrate that models pre-trained without legal data underperform in the task of negation scope resolution. Our experiments, using language models exclusively fine-tuned on domains like literary texts and medical data, yield inferior results compared to the outcomes documented in prior cross-domain experiments. We release a new set of annotated court decisions in German, French, and Italian and use it to improve negation scope resolution in both zero-shot and multilingual settings. We achieve token-level F1-scores of up to 86.7\% in our zero-shot cross-lingual experiments, where the models are trained on two languages of our legal datasets and evaluated on the third. Our multilingual experiments, where the models were trained on all available negation data and evaluated on our legal datasets, resulted in F1-scores of up to 91.1\%. 
\end{abstract}

\section{Introduction}

Negation scope resolution is an important research problem in the field of Natural Language Processing (NLP). It describes the detection of words that are affected by a negation cue (e.g. no or not) in a sentence, which is important for understanding its true meaning. Although this task is far from trivial, deep learning approaches have shown promising results \citep{khandelwal_negbert_2020, shaitarova_cross-lingual_2020, shaitarova_negation_2021}. 

\begin{figure}[t]
    \includegraphics[width=\columnwidth]{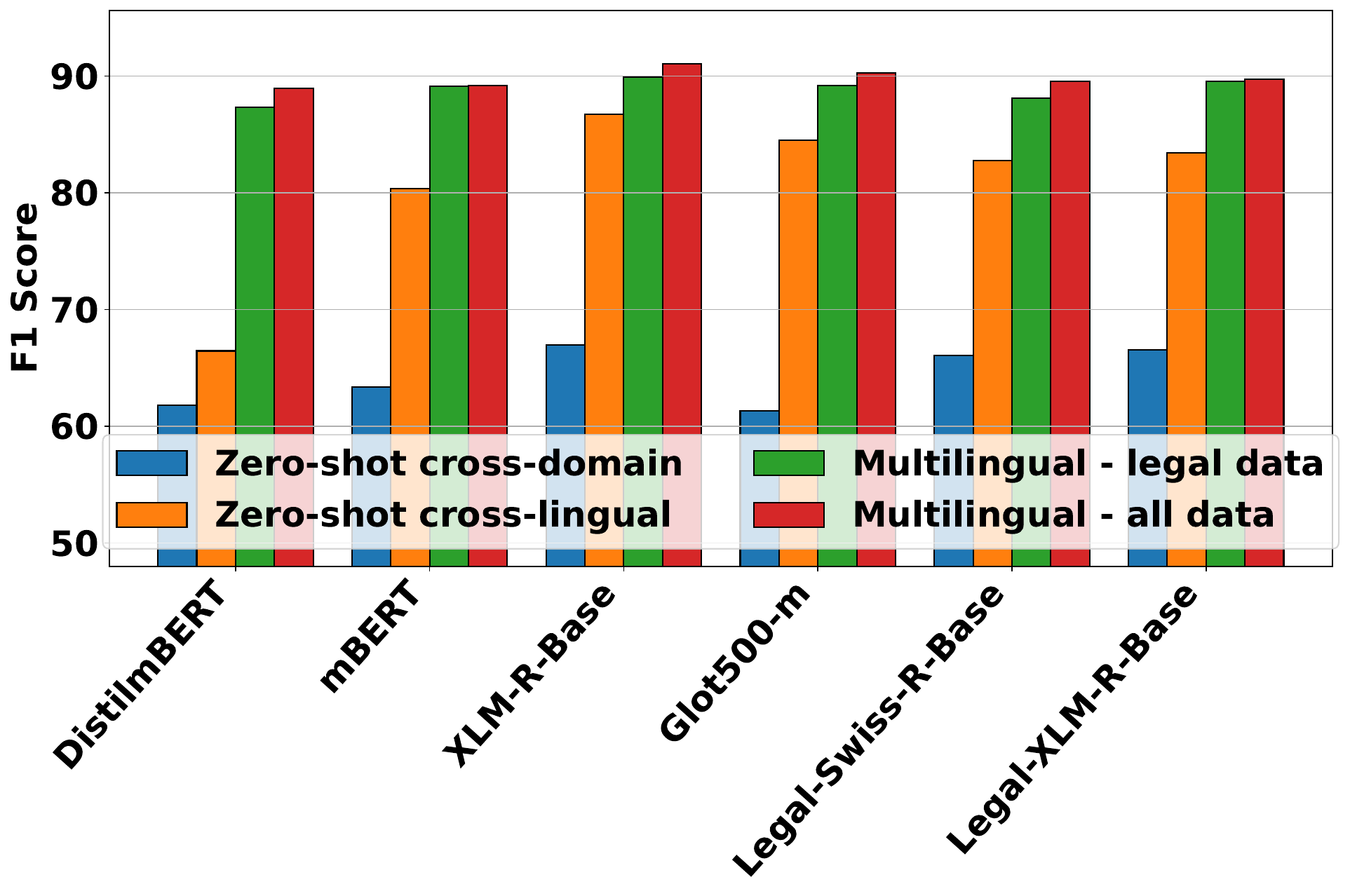}
  \caption{Results over main experiments from select models. For all results see \Cref{sec:all_results}.}
\end{figure}

As with many NLP tasks, the largest amount of annotated data is available in English.\footnote{\cite{Mie2016Language} analyzed all ACL conference proceedings from 2004, 2008, 2012, and 2016 and found that between 58\% and 69\% of papers only evaluated in English.} Multilingual datasets are less common and often not easily accessible. For example, on the \href{https://huggingface.co/}{huggingface hub}, hosting most important open-source datasets, 4559 datasets are tagged as English. The next most common language is Chinese with 10 times fewer datasets for a total of 469.\footnote{Numbers extracted from \href{https://huggingface.co/datasets}{https://huggingface.co/ datasets} on 13.08.2023.}\ In addition, much of the work conducted in the area of negation scope resolution has been done in the medical domain in order to automatically process clinical reports and discharge summaries \cite{szarvas2008bioscope}. Other datasets consist of literary texts \cite{morante2012sherlock} or more informal data such as online reviews \cite{konstantinova2012sfu}. The legal domain differs from all of the above in that it is often very complex (i.e., legalese) and uses highly specific vocabulary and knowledge that is not common outside the legal domain \cite{friedrich2021complexitylaw, ruhl2017complexitylaw2}. This poses a challenge to any model tackling tasks in the legal domain. While a large amount of legal data is publicly available and has been annotated for various tasks \cite{multieurlexLegaldata,rasiah_scale_2023,niklaus_swiss-judgment-prediction_2021,niklaus_lextreme_2023,brugger_multilegalsbd_2023,niklaus_multilegalpile_2023,chalkidis_lexglue_2022}, \emph{inter alia}, to the best of our knowledge there exists no legal negation corpus.

We annotate four new datasets containing legal judgments from Swiss and German courts in German, French and Italian for negation cues and scopes. We find that these legal documents contain on average longer sentences as well as longer annotated negation scopes, compared to existing datasets.
Our experiments show that the legal domain poses a significant challenge to models attempting negation scope resolution. The results achieved by models pre-trained in different domains and evaluated on legal data are lower than those seen in other cross-corpus experiments \citep{khandelwal_negbert_2020, shaitarova_negation_2021}. Using our newly annotated datasets, we can improve these results. We conduct experiments where the models are fine-tuned on two languages of the legal data and evaluated on the third. In these zero-shot cross-lingual experiments, our models achieve higher F1-scores than the models pre-trained only on different domains. By training on all available data, we are able to further improve these results, achieving F1-scores around 90\% for our multilingual experiments. Our results provide an interesting insight into how even smaller datasets can make a valuable contribution to improving the performance of \acp{lm} on a specific downstream task such as negation scope resolution.

\subsection*{Contributions}
The contributions of this paper are three-fold:
\begin{compactitem}
    \item We annotate new datasets of legal documents for negation in German, French, and Italian each containing around 1000 sentences.
    \item We train and evaluate models on the task of negation scope resolution on the newly annotated datasets to provide a reference point and achieve token-level F1-scores in the mid eighties for cross-lingual zero-shot experiments and up to 91\% in multilingual experiments. 
    \item We publicly release the annotation guidelines, the data, the models and the experimentation code as resources and for reproducibility.\footnote{The annotation guidelines as well as the code to fine-tune our models can be found on GitHub:  \href{https://github.com/RamonaChristen/Multilingual_Negation_Scope_Resolution_on_Legal_Data}{https://github.com/RamonaChristen/Multilingual\_Negation\_ Scope\_Resolution\_on\_Legal\_Data}. Our best model (\href{https://huggingface.co/rcds/neg-xlm-roberta-base}{https://huggingface.co/rcds/neg-xlm-roberta-base}) and dataset (\href{https://huggingface.co/rcds/ MulitLegalNeg}{https://huggingface.co/rcds/MulitLegalNeg}) are published on huggingface.}
\end{compactitem}

\section{Related Work}
Different approaches have been used to address the issue of negation detection and negation scope resolution. Early research focused mainly on rule-based approaches. NegEx, a simple regular expression algorithm developed by \citet{chapman2001}, was successfully able to identify negations in the medical domain. \citet{morante2008} first took a machine learning approach to negation scope resolution. They used two memory-based classifiers, one to identify the negation cue in a sentence, and one to identify the scope of the negation. On the negation scope resolution task, they achieved an F1-score of 81\% on the BioScope corpus \cite{szarvas2008bioscope}. These results were later surpassed by \citet{fancellu2017neural}, achieving an F1-score of 92\% by using neural networks for scope detection. 
\citet{khandelwal_negbert_2020} achieved the best results on the BioScope corpus, as well as on two other publicly available negation corpora, the SFU Review Corpus \cite{konstantinova2012sfu} and the ConanDoyle-neg corpus \cite{morante2012sherlock}.\ Their NegBERT model uses \ac{bert} \cite{devlin2018bert} and applies a transfer learning approach for negation detection and scope resolution.

Only a limited amount of work has been conducted on negation scope resolution across different languages. \citet{fancellu2018multilingual} developed a cross-lingual system, trained on English data and tested on a Chinese corpus. By employing cross-lingual universal dependencies in English they were able to achieve an F1-score of 72\% on the Chinese data. \citet{shaitarova_cross-lingual_2020} investigated cross-lingual zero-shot negation scope resolution between English, Spanish, and French. They built on NegBERT but used the multilingual BERT (mBERT) model. \citet{shaitarova_negation_2021} built on this using NegBERT with mBERT and XLM-R\textsubscript{Large} \cite{conneau_unsupervised_2020}, and were able to achieve a token-level F1-score of 87\% on zero-shot transfer from Spanish to Russian.

The sparse amount of cross-lingual research can be explained by the lack of annotated data in languages other than English. There are few corpora annotated with negations in German and Italian \cite{jimenez2020corpora}. The only German corpus annotated for negation and speculation contains medical data and clinical notes \cite{cotikGermanCorpus}. However, the corpus is not publicly available and no annotation guidelines have been published. For Italian, \citet{altunaItalian} presented a framework for the annotation of negations and applied it to a corpus of news articles and tweets, parts of which are publicly available. In French, \citet{dallouxFrench} annotated a medical corpus, available on request. To our knowledge, no legal corpus annotated with negations currently exists.

\section{Data}
\subsection{Legal Data}
We use court decisions in our legal datasets, also often referred to as judgments. The judgments form German courts were collected from \emph{Bayern.Recht}\footnote{\href{https://www.gesetze-bayern.de/}{https://www.gesetze-bayern.de/}} and include a variety of legal domains and structures \citep{gerDataset}. The Swiss court decisions in French, Italian, and German (CH) were collected from the \ac{fscs}.
The \ac{fscs} is the highest legal authority in Switzerland and oversees federal criminal, administrative, patent, and cantonal courts.

Judgments published by the \ac{fscs} usually consist of four sections: 1) The introduction gives information about the date, chamber, involved judge(s) and parties, and the topic of the court decision. 2) The facts outline the important case information. 3) The considerations form the basis for the final ruling by providing relevant case law  and other cited rulings. 4) The rulings gives the final decision made by the court.

\subsection{Datasets}
We annotated four new datasets in three languages for negation cues and scopes, and standardized the existing French and English datasets to make them more accessible. Our datasets consist of publicly available legal judgments from Swiss and German courts. Since negation scope resolution is a sentence-level task, we first split the data into sentences using sentence boundary annotations. The French (fr) and Italian (it) datasets consist of a subset of Swiss court decisions from the Swiss-Judgment-Prediction (SJP) dataset \citep{niklausJudgmentCorpus} and the \href{https://huggingface.co/datasets/joelito/Multi_Legal_Pile}{Multi-Legal-Pile} \cite{niklaus_multilegalpile_2023} which were annotated for sentence spans by \citet{brugger_multilegalsbd_2023}. The main German data (de (DE)) is a subset of judgments from German courts collected by \citet{gerDataset}. 
Only judgments were included in our dataset because they include a variety of sources and legal areas, they also have a higher density of negation cues compared to other legal texts. To validate that the negation scope prediction also works on German court data from Switzerland, we curated a small dataset of German-Swiss court decisions (de (CH)) that is also a subset of the SJP corpus. We separated each dataset into a train (70\%), test (20\%), and validation (10\%) split.

To ensure that sufficient negation data is available in each dataset, a negation score was assigned to each document based on a simple word search for the most common negation words in each language. The documents with the highest negation scores were then selected to be annotated. \Cref{tab:datasetStatistics} shows the amount of data and the distribution of negations for the newly created datasets in comparison to the existing datasets in English and French. Our datasets contain a slightly higher ratio of negated sentences compared to the other datasets. This can be attributed to the nature of legal data and our pre-selection procedure. Because we annotated only a subset of an existing corpus we were able to exclude documents without or only few negations while other corpora like ConanDoyle-neg and SFU annotated complete existing datasets or stories.

\begin{table}[ht]
    \centering
    \small
    \resizebox{\columnwidth}{!}{
    \begin{tabular}{clccc}
    \toprule
        & \textbf{Dataset} & \textbf{Total} & \textbf{Negated} & \%neg \\
        \midrule
         \parbox[t]{2mm}{\multirow{4}{*}{\rotatebox[origin=c]{90}{legal}}} 
            & fr & 1059 & 382 & 36.07\\
            & it & 1001 & 418 & 41.76\\
            & de (DE)& 1098 & 454 & 41.35\\
            & de (CH) & 208 & 112 & 53.85\\
        \midrule
         \parbox[t]{2mm}{\multirow{4}{*}{\rotatebox[origin=c]{90}{external}}} & SFU & 17672 & 3528 & 19.96\\
         & BioScope & 14700 & 2095 & 14.25\\
         & ConanDoyle-neg & 5714 & 1421 & 24.87\\
         & Dalloux & 11032 & 1817 & 16.47\\
        \bottomrule
    \end{tabular}
    }
    \caption{Total number of sentences, and number and percentage of sentences containing at least one negation.}
    \label{tab:datasetStatistics}
\end{table}

Annotations were done by native-language human annotators using the tool  \href{https://prodi.gy/}{Prodigy}. All annotators are university students but not part of a legal study program. The annotations were cross-checked by one annotator, who has a linguistic background, with the help of an online translator to ensure that they adhere to the annotation guidelines and are consistent across all three languages. The annotation guidelines are based on existing guidelines for the English datasets, and have been extended to cover all three languages included in our data, as well as the characteristics of the legal domain. Key guidelines are summarized below.

\paragraph{Negation Cues} Cues were not annotated as part of the negation scope following the annotation guidelines for the ConanDoyle-neg corpus \cite{morante2011annotation}. We excluded affixal cues\footnote{Affixal cues are cues within a word such as \ul{im}possible} in our annotations and kept all annotations to the word as the level of the minimal syntactic unit.  

\paragraph{Multiple negations} Annotators were instructed to annotate one negation per sentence. Sentences with multiple negations were duplicated before annotation based on the most common negation cues. To ensure that the same cue was not annotated twice, duplicates were displayed next to each other in the annotation tool to allow annotators to see which clues had yet to be annotated.

\paragraph{Maxiumum scope strategy} As with BioScope, we used a maximum scope strategy. This means that the scope extends to the largest possible unit. If a negated clause has subordinate clauses providing additional information to the clause, the scope extends over the negated clause and all of its subordinate clauses, as illustrated in example 1. This sentence structure is very common in our set of legal data. In all following examples we mark the cue in \textbf{bold} and \ul{underline} the scope. We provide an English translation for clarity.
 \begin{enumerate}
     \item[1] \ul{Vorliegend ginge es} \textbf{nicht} \ul{darum, dass ein Arbeitgeber \"uber Fristen oder Pflichten nicht aufgekl\"art habe, somit eine blosse Unt\"atigkeit des Arbeitgebers} [...]
     \label{itemzero}
     \item [\textit{EN:}] \textit{\ul{In the present case, it was} \textbf{not} \ul{a matter of an employer not having provided information about deadlines or obligations, thus a mere inactivity on the part of the employer} [...].}
 \end{enumerate}

\paragraph{Case citations} Our dataset contains two main types of citations: inline citations and parenthesized citations. Inline citations, as in example 2 were annotated as part of the scope, while parenthesized citations, as in example 3, were excluded from the negation scope.
\begin{enumerate}
   \item[2] Da \ul{der Kl\"ager} \textbf{kein} \ul{\"ahnlicher leitender Angestellter i.S.d 14 Abs. 2Satz 2 KSchG ist} [...]
 \label{itemone}
    \item[\textit{EN:}] \textit{Since \ul{the plaintiff is} \textbf{not} \ul{a similar executive employee in the sense of 14 Abs. 2Satz 2 KSchG}} [...]
   \item[3]\ul{Seit dem 06.02.2017 ist der Kl\"ager im Handelsregister} \textbf{nicht mehr} \ul{als Gesch\"aftsf\"uhrer eingetragen} (vgl. Auszug aus dem Handelsregister in Anlage K9, Bl 75 ff. d.A).
   \item[\textit{EN:}] \textit{\ul{Since 06.02.2017 the plaintiff is} \textbf{no longer} \ul{registered in the commercial register as managing director} (see extract from the commercial register in annex K9, Bl 75 ff. d.A)}
 \label{itemtwo}
\end{enumerate}

\paragraph{Punctuation} Punctuation marks, such as periods or exclamation points, were excluded from the scope, unless the scope spans multiple clauses separated by commas.

\Cref{tab:sentence_length} shows the average number of tokens in a sentence for all datasets, as well as the average length of the annotated scopes as a ratio between annotated and not annotated tokens. On average, the sentences in our legal datasets are slightly longer than in other datasets. Furthermore, the mean length of the annotated scopes in our data is higher than in all other datasets. For de (DE), more than 50\% of tokens were annotated as scope, which is around twice as much as with the biomedical, literary, and review corpora. This is due to the legal domain's sentence structure and our annotation guidelines, which include the subject in the scope. Additionally, nested sentences with multiple subordinate clauses are common in our dataset. This, combined with our maximum scope strategy, leads to longer scopes compared to other datasets.

\begin{table}[ht]
    \centering
    \small
    \resizebox{\columnwidth}{!}{
    \begin{tabular}{llcc}
    \toprule
        & \textbf{Dataset} & \textbf{Sentence} & \textbf{Scope}\\
        \midrule
        \parbox[t]{2mm}{\multirow{4}{*}{\rotatebox[origin=c]{90}{legal}}} 
         & fr &  48.52 & 37.96\% \\
         & it & 40.84 & 30.17\%\\
         & de (DE) & 31.14 & 50.18\%\\
         & de (CH) & 27.65 & 36.03\%\\
         \midrule
         \parbox[t]{2mm}{\multirow{4}{*}{\rotatebox[origin=c]{90}{external}}} &
         SFU & 24.46 & 21.87\%\\
         & BioScope &  28.49 & 25.91\%\\
         & ConanDoyle-neg & 22.11 & 32.37\%\\
         & Dalloux & 25.96 & 19.82\%\\
         \bottomrule
    \end{tabular}
    }
    \caption{The average number of tokens per sentence.\ Scopes are shown as a percentage of negated tokens.}
    \label{tab:sentence_length}
\end{table}

\section{Experimental setup}
We performed experiments to assess negation scope resolution model performance on our multilingual legal data. We integrated the NegBERT architecture \cite{khandelwal_negbert_2020}, successful in this task on prior datasets, with various pre-trained multilingual \acp{lm} outlined in \Cref{tab:models}. We ran each experiment five times with different random seeds and report the mean token-level F1-score averaged over random seeds, together with the standard deviation. All experiments were conducted with the same hyperparameters for all models, optimized with a search over learning rate (5e-7, 1e-6, 3e-6, 1e-5, 3e-5, 5e-5) and batch size (4, 8, 16, 32, 64, and 128). We optimized the hyperparameters for mBERT and XLM-R and concluded that the best results can be achieved with an initial learning rate of 1e-5 and a batch size of 16. To avoid overfitting, we used early stopping with patience set to 8 as a compromise between the patience of 6 used in the original NegBERT experiments \citep{khandelwal_negbert_2020} and 9 used in the multilingual experiments of \citet{shaitarova_negation_2021}. We extended the maximum input length to 252 tokens for our data. Experiments ran on an NVIDIA A100 GPU via Google Colab, totaling around 105 hours of training time.



\begin{table*}[ht]
  \centering
  \resizebox{\textwidth}{!}{
  \begin{tabular}{llrrrrrr}
    \toprule
    \textbf{Model} & \textbf{Source} & \textbf{InLen} & \textbf{Params} & \textbf{Vocab} &  \textbf{NumTokens} &\textbf{Corpus} & \textbf{Langs} \\
    \midrule
    DistilmBERT                             & \citet{sanh_distilbert_2020}          & 512   & 134M      & 120K &  n/a         & Wikipedia               & 104\\
    mBERT                                   & \citet{devlin2018bert}                & 512   & 177K      & 120K &  n/a         & Wikipedia               & 104 \\

    XLM-R\textsubscript{Base/Large}         & \citet{conneau_unsupervised_2020}     & 512   & 278M/560M & 250K & 6'291B               & 2.5TB CC100              & 100\\
    \midrule
    Glot500-m                               & \citet{glot500}                       & 512   & 395M      & 401K & 94B           & glot500-c                & 511\\
    \midrule
    Legal-Swiss-R\textsubscript{Base/Large} & \citet{rasiah_scale_2023}             & 512   & 184M/435M & 128K & 262B/131B            & CH Rulings/Legislation   & 3\\
    Legal-XLM-R\textsubscript{Base/Large}   & \citet{niklaus_multilegalpile_2023}   & 512  & 184M/435M & 128K &  262B/131B           &CH Rulings/Legislation   & 3\\
    \bottomrule
  \end{tabular}
  }
   \caption{Model stats. InLen: max input length during pre-training. Params: total parameter count. NumTokens: Batch size $\times$ Steps $\times$ InLen}
  \label{tab:models}
\end{table*}
Firstly, we evaluated \nameref{para:chatgpt} in zero- and few-shot experiments to interpret the results of a non-fine-tuned model in the negation scope resolution task. For all subsequent experiments, we used the NegBERT architecture. In the first NegBERT experiment, models were fine-tuned on all existing French and English datasets and evaluated on our new legal datasets, representing a \nameref{para:crossDomain}. For a second series of zero-shot experiments, we attempted a \nameref{para:zero_cross_lingual} within our legal datasets. In each cross-lingual experiment, models were trained on two dataset languages and evaluated on the third. We also executed \nameref{para:multilingual} using our datasets and all available data.

\section{Results}
\paragraph{ChatGPT}
\label{para:chatgpt}
We evaluated the performance of ChatGPT-3.5 \cite{brown_language_2020}, one of the leading \acp{lm}, in the task of negation scope resolution on our legal datasets. Other researchers have found that ChatGPT performs well on simple annotation tasks such as text classification \cite{gilardi2023chatgpt}. To analyze ChatGPT's understanding of negation scopes, we conducted a small test over the chat interface (See \Cref{sec:chatGPT}) which showed that it was able to correctly identify the negation scope of a simple German sentence. For the same request with an example sentence from our legal dataset, ChatGPT was not able to accurately identify the negation scope. To evaluate the performance on the whole dataset, we used the ChatGPT API with `gpt-3.5-turbo-16k' to accommodate longer inputs. We set the temperature to 0 to reduce randomness and receive a coherent output in json format. Similar to the experiments with the NegBERT architecture, we gave the sentence as well as the negation cues as input and prompted ChatGPT to return the sentence annotated for negation scopes. In a zero-shot experiment, we did not give any annotated examples and only provided a short definition of negation scopes. The results show that ChatGPT's performance on our datasets is subpar (\Cref{tab:gptResults}). In an effort to increase the performance, we conducted some few-shot experiments where 1, 5 or 10 examples of annotated sentences were provided with the prompt, but it did not lead to improvement. The results of the 1-shot experiments averaged lower than the 0-shot experiments. Overall the standard deviation is very high which can be explained by the fact that a random set of annotated examples was selected for each of the five experiment runs. Overall we can conclude, that ChatGPT is currently not suited to solve negation scope resolution in the legal domain without fine-tuning.
\begin{table*}[ht]
    \centering
    \resizebox{\textwidth}{!}{
    \begin{tabular}{llllll}
\toprule
\textbf{Test Dataset} &     \textbf{0-shot}          &     \textbf{1-shot}           &    \textbf{5-shot}            &  \textbf{10-shot} & \Centerstack{Mean F1 \\by Dataset} \\
\midrule
fr           &  $13.00_{\pm 2.1}$ &  $16.63_{\pm 10.3}$ &   $14.90_{\pm 7.5}$ &  $22.53_{\pm 10.7}$ &  $16.77_{\pm8.5}$\\
it           &  $25.11_{\pm 1.5}$ &   $18.22_{\pm 6.5}$ &   $31.07_{\pm 7.1}$ &   $26.10_{\pm 3.8}$ & $25.12_{\pm6.7}$\\
de (DE)          &  $16.47_{\pm 2.6}$ &   $22.45_{\pm 9.1}$ &   $17.34_{\pm 2.7}$ &  $24.48_{\pm 10.7}$ & $20.18_{\pm7.5}$\\
de (CH)           &  $32.91_{\pm 7.9}$ &   $21.20_{\pm 5.8}$ &  $36.89_{\pm 18.6}$ &  $19.83_{\pm 10.3}$ & $27.71_{\pm13.1}$\\
\bottomrule
\textbf{Mean F1 by experiment} & $21.87_{\pm8.9}$ & $19.62_{\pm7.8}$ & $25.05_{\pm13.6}$ & $23.23_{\pm8.9}$ & \\
\bottomrule
    \end{tabular}
    }
    \caption{Results for zero- and few-shot experiments conducted over the ChatGPT API.}
    \label{tab:gptResults}
\end{table*}

\paragraph{Zero-shot cross-domain transfer}
\label{para:crossDomain}
The results for our zero-shot cross-domain transfer experiments are presented in \Cref{tab:zeroExternal}. The best results over all datasets were achieved by the Legal-XLM-R\textsubscript{Large} model, scoring an F1-score of 71.6\%. Overall, the \acp{lm} pre-trained on legal data demonstrated a 4-percentage point advantage, with a mean F1 of 68.3\% averaged over all four legal models, compared to the other models pre-trained on different domains. Furthermore, we notice that the standard deviation for the experiments conducted with the \acp{lm} pre-trained on legal data is higher compared to the other models. A possible explanation is that pre-training on legal data improved negation predictions in some areas but adversely affected others, likely due to bias in the legal models, thereby increasing standard deviations across experiments. Generally, cross-domain transfer to the legal domain is less successful than other zero-shot experiments across languages and domains (i.e., \citet{shaitarova_cross-lingual_2020, khandelwal_negbert_2020}). This suggests that transferring from non-legal to legal domains is challenging.

\begin{table*}[ht]
    \centering
    \resizebox{\textwidth}{!}{
\begin{tabular}{l cccc l}\hline
\diagbox{Model}{Test Dataset}
   &         fr &          it &           de (DE) &           de (CH) & \Centerstack {Mean F1 \\ by Model}\\
\midrule
    DistilmBERT                         & $61.43_{\pm 1.9}$    & $63.40_{\pm 2.6}$ & $63.50_{\pm 4.3}$ & $58.78_{\pm 4.5}$ & $61.78_{\pm 3.8}$\\
    mBERT                               & $66.39_{\pm 2.1}$    & $68.49_{\pm 0.8}$ & $64.17_{\pm 3.1}$ & $54.31_{\pm 4.8}$ & $63.34_{\pm 6.2}$ \\
    XLM-R\textsubscript{Base}           & $66.80_{\pm1.9}$     & $71.40_{\pm0.8}$ & $67.29_{\pm3.7}$ & $62.44_{\pm2.9}$ & $66.98 _{\pm 4.0}$\\
    XLM-R\textsubscript{Large}          & $72.30_{\pm 2.0}$    & $70.30_{\pm 0.9}$ & $73.81 _{\pm 4.2}$ & $\textbf{63.72}_{\pm 4.6}$ & $70.03_{\pm5.0}$\\
    \midrule
    Glot500-m                           & $63.78_{\pm 0.8}$    & $65.54_{\pm 1.1}$ & $61.38_{\pm 4.0}$ & $54.51_{\pm 2.5}$ & $61.30_{\pm 4.9}$\\
    \midrule
    Legal-Swiss-R\textsubscript{Base}   & $69.48_{\pm 2.3}$    & $68.64_{\pm 1.0}$ & $71.81_{\pm 3.8}$ & $54.26_{\pm 4.9}$ & $66.05_{\pm 7.7}$ \\
    Legal-Swiss-R\textsubscript{Large}  & $\textbf{74.66}_{\pm 2.4}$ & $72.68_{\pm 1.5}$ & $\textbf{76.5}_{\pm 1.6}$ & $51.75_{\pm 6.6}$ &$ 68.89_{\pm 10.8}$\\
    Legal-XLM-R\textsubscript{Base}     & $71.50_{\pm 3.1}$    & $71.48_{\pm 2.2}$ & $71.35_{\pm 5.4}$ & $51.93_{\pm 3.5}$ & $66.57_{\pm 9.3}$\\
    Legal-XLM-R\textsubscript{Large}    & $74.52_{\pm 2.1}$    & $\textbf{74.48}_{\pm 3.3}$& $76.06_{\pm 3.3}$ & $61.30_{\pm 8.9}$ & $\textbf{71.59}_{\pm 7.7}$\\
    \midrule
    \textit{ChatGPT}                    & $13.00_{\pm2.1}$      & $25.11_{\pm1.5}$ & $16.47_{\pm2.6}$ & $32.91_{\pm7.9}$ & $21.87_{\pm8.9}$ \\
    \midrule
Mean F1 by Dataset &  $68.99_{\pm 4.9}$ & $\textbf{69.60}_{\pm 3.7}$ & $69.54_{\pm 6.4}$ & $57.00_{\pm 6.4}$ & $66.28_{\pm 7.6}$  \\    

\bottomrule
\end{tabular}}
    \caption{Cross-domain zero-shot results from existing datasets to our new legal datasets. All models except for ChatGPT were pre-trained on all external datasets, ChatGPT did not receive any training data. The bottom right entry shows the average across all datasets and models except ChatGPT.}
    \label{tab:zeroExternal}
\end{table*}

\paragraph{Zero-shot cross-lingual transfer}\label{para:zero_cross_lingual}
\Cref{tab:zeroLegal} presents the results of our zero-shot cross-lingual experiments conducted with only our legal data. Although these datasets are considerably smaller than the existing English and French datasets, we were able to increase the F1-score by an average of 15.6\% across all models and datasets. The legal models still performed well in these experiments, but they no longer showed an advantage over the other \acp{lm}. XLM-R\textsubscript{Base} achieved the best results. All models, except for DistilmBERT, performed significantly better than in the previous experiment across all datasets.\ DistilmBERT performed worse on the German datasets than in the previous experiment. One explanation for this might be that DistilmBERT is the only cased model used in our experiments.\ While cased models usually outperform uncased models, this does not seem to apply to cross-lingual experiments.\ Similar results were found by \citet{uncasedLearningTransfer}, who conducted cross-lingual reading comprehension experiments from English to Czech and found that the uncased models outperformed the cased models in these experiments.\ They theorized that the overlap of sub-words is larger between English and Czech for uncased models because they disregard diacritical marks, which are common in Czech. A similar argument could be made for the cross-lingual transfer between Italian, French, and German because German includes a lot of casing information while Italian and French do not. 

\begin{table*}[ht]
    \centering
    \resizebox{\textwidth}{!}{
    \begin{tabular}{lccccl}
    \toprule
    \diagbox{Model}{Test Dataset} &    fr &  it &   de (DE) & de (CH) & \Centerstack{Mean F1 \\by Model} \\
    \midrule
        DistilmBERT                         & $79.56_{\pm 1.0}$ & $74.94_{\pm 1.7}$ & $58.74_{\pm 9.6}$ & $52.59_{\pm 11.3}$ &  $66.46_{\pm 13.3}$\\
        mBERT                               & $87.22_{\pm 1.6}$ & $81.94_{\pm 1.3}$ & $81.39_{\pm 3.6}$ &  $70.78_{\pm 6.7}$ & $80.33_{\pm 7.1}$\\
        XLM-R\textsubscript{Base}           & $88.70_{\pm 0.8}$ & $\textbf{86.43}_{\pm 2.2}$ & $88.00_{\pm 1.9}$ & $\textbf{83.71}_{\pm 4.8}$ & $\textbf{86.71}_{\pm 3.3}$\\
        XLM-R\textsubscript{Large}  & $\textbf{90.55}_{\pm 0.9}$ & $84.93_{\pm 1.7}$ & $\textbf{91.36}_{\pm 0.8}$ &  $76.65_{\pm 4.5}$ & $85.87_{\pm 6.4}$\\
        \midrule
        Glot500-m                           & $86.77_{\pm 2.3}$ & $83.41_{\pm 1.3}$ & $90.10_{\pm 2.0}$ &  $77.73_{\pm 4.6}$ & $84.50_{\pm 5.4}$ \\
        \midrule
        Legal-Swiss-R\textsubscript{Base}   & $87.42_{\pm 1.2}$ & $84.54_{\pm 1.6}$ & $88.24_{\pm 1.0}$ &  $70.95_{\pm 3.6}$ & $82.79_{\pm 7.4}$\\
        Legal-Swiss-R\textsubscript{Large}  & $84.63_{\pm 1.0}$ & $83.88_{\pm 1.9}$ & $88.47_{\pm 3.9}$ &  $70.33_{\pm 6.0}$ & $81.83_{\pm 7.8}$\\
        Legal-XLM-R\textsubscript{Base}     & $86.40_{\pm 2.1}$ & $83.28_{\pm 1.4}$& $89.56_{\pm 2.5}$ &  $74.52_{\pm 8.0}$ & $83.44_{\pm 7.0}$\\
        Legal-XLM-R\textsubscript{Large}    & $85.51_{\pm 1.7}$ & $85.76_{\pm 0.3}$ & $89.58_{\pm 1.8}$ &  $80.16_{\pm 4.0}$ & $85.25_{\pm 4.1}$\\
    \midrule
        Mean F1 by dataset & $\textbf{86.31}_{\pm 3.2}$ &$ 83.23_{\pm 3.5}$ & $85.05_{\pm 10.4}$ & $73.05_{\pm 10.3}$ & $81.91_{\pm 9.3}$\\
               
    \bottomrule
\end{tabular}}
    \caption{Multilingual zero-shot experiments within our legal datasets. Each column represents a different set of test and train data where the test data includes all legal datasets in languages that are not the language of the test dataset i.e. models evaluated on fr were trained with it and de (DE,CH).}
    \label{tab:zeroLegal}
\end{table*}

\paragraph{Multilingual experiments}
\label{para:multilingual}
The best results for negation scope resolution on our legal datasets were achieved by training our models on the entirety of the available data (\Cref{tab:Mulilingual}). This multilingual approach achieved an average F1-score of 90\% across all models and datasets and outperformed all of the previous setups. This indicates that a relatively small amount of training data in the domain and language of the test dataset can significantly improve the performance of a \ac{lm}. It is also notable that there seems to be no substantial difference in the performance of the different \acp{lm} in this experiment, with a standard deviation of only $\pm$ 3.6 over all models and datasets. Although DistilmBERT obtained the lowest scores in this experiment, its performance is not significantly inferior to that of the mBERT model. This could be attributed to the fact that the training data also included German examples which might have mitigated the advantage of the uncased models with regard to shared vocabulary. We also conducted multilingual experiments only using our new datasets which achieved very similar results with an overall F1-score of  $89.1_{\pm4}$ (see \Cref{sec:multilingual_legal_results}).

\begin{table*}[ht]
    \centering
    \resizebox{\textwidth}{!}{
    \begin{tabular}{llllll}
        \toprule
         \diagbox{Model}{Test Dataset} &           fr &           it &           de (DE) &           de (CH)  &       \Centerstack{Mean F1\\ by Model}\\

        \midrule
            DistilmBERT & $87.54_{\pm 0.6}$ & $82.90_{\pm 1.3}$ & $94.63_{\pm 0.5}$ & $90.77_{\pm 1.2}$ & $88.96_{\pm 4.5}$\\
            mBERT & $89.98_{\pm 2.1}$ & $83.72_{\pm 1.0}$ & $95.21_{\pm 0.5}$ & $87.83_{\pm 1.0}$ & $89.10_{\pm 4.4}$  \\
            XLM-R\textsubscript{Base} & $\textbf{91.31}_{\pm 1.2}$ & $88.81_{\pm 1.1}$ & $94.74_{\pm 0.7}$ & $89.39_{\pm 1.8}$ & $\textbf{91.06}_{\pm 2.6}$\\
            XLM-R\textsubscript{Large} & $90.77_{\pm 1.8}$ & $87.44_{\pm 0.5}$ & $93.40_{\pm 1.1}$ & $90.20_{\pm 3.9}$ & $90.45_{\pm 3.0}$\\
            \midrule
            Glot500-m & $89.65_{\pm 1.0}$ & $85.54_{\pm 2.3}$ & $94.94_{\pm 0.7}$ & $\textbf{91.00}_{\pm 2.7}$ & $90.28_{\pm 3.8}$\\
            \midrule
            Legal-Swiss-R\textsubscript{Base} & $89.08_{\pm 1.6}$ & $87.40_{\pm 1.9}$ & $94.60_{\pm 1.0}$ & $87.02_{\pm 1.5}$ & $89.52_{\pm 3.4}$\\
            Legal-Swiss-R\textsubscript{Large} & $89.07_{\pm 1.4}$ & $86.72_{\pm 1.5}$ & $\textbf{95.94}_{\pm 0.2}$ & $89.39_{\pm 0.9}$ & $90.28_{\pm 3.7}$\\
            Legal-XLM-R\textsubscript{Base} & $90.71_{\pm 0.5}$ & $86.67_{\pm 0.5}$ & $95.41_{\pm 0.7}$ & $86.17_{\pm 2.4}$ & $89.74_{\pm 4.0}$\\
            Legal-XLM-R\textsubscript{Large} & $90.75_{\pm 1.4}$ & $\textbf{89.46}_{\pm 0.8}$ & $93.87_{\pm 0.8}$ & $89.18_{\pm 1.0}$ & $90.82_{\pm 2.1}$\\
        \midrule 
            Mean F1 by Dataset & $89.87_{\pm 1.2}$   & $86.52_{\pm 2.4}$ & $\textbf{94.74}_{\pm 1.0}$ & $88.99_{\pm 2.4}$ & $90.03_{\pm 3.6}$\\
        \bottomrule
        \end{tabular}
}
    \caption{Results from multilingual experiments over all available data.}
    \label{tab:Mulilingual}
\end{table*}

\subsection{Error analysis}
We investigated the length of the predicted negation scopes as well as random samples of the predictions on the French and German test data to identify some common error cases.

\paragraph{Predicted scope length}
As expected, our cross-domain zero-shot experiments without legal training data achieved the lowest F1-scores overall. This can mostly be attributed to the differences in annotation for each dataset, as well as the different domains.\ Although the external corpora included French data, this did not improve the performance on the French dataset compared to the other legal datasets. A possible reason is that the subject was not annotated as part of the scope in the Dalloux dataset opposed to the French legal dataset. 

\begin{figure}[ht]
    \centering
    \includegraphics[width=\columnwidth]{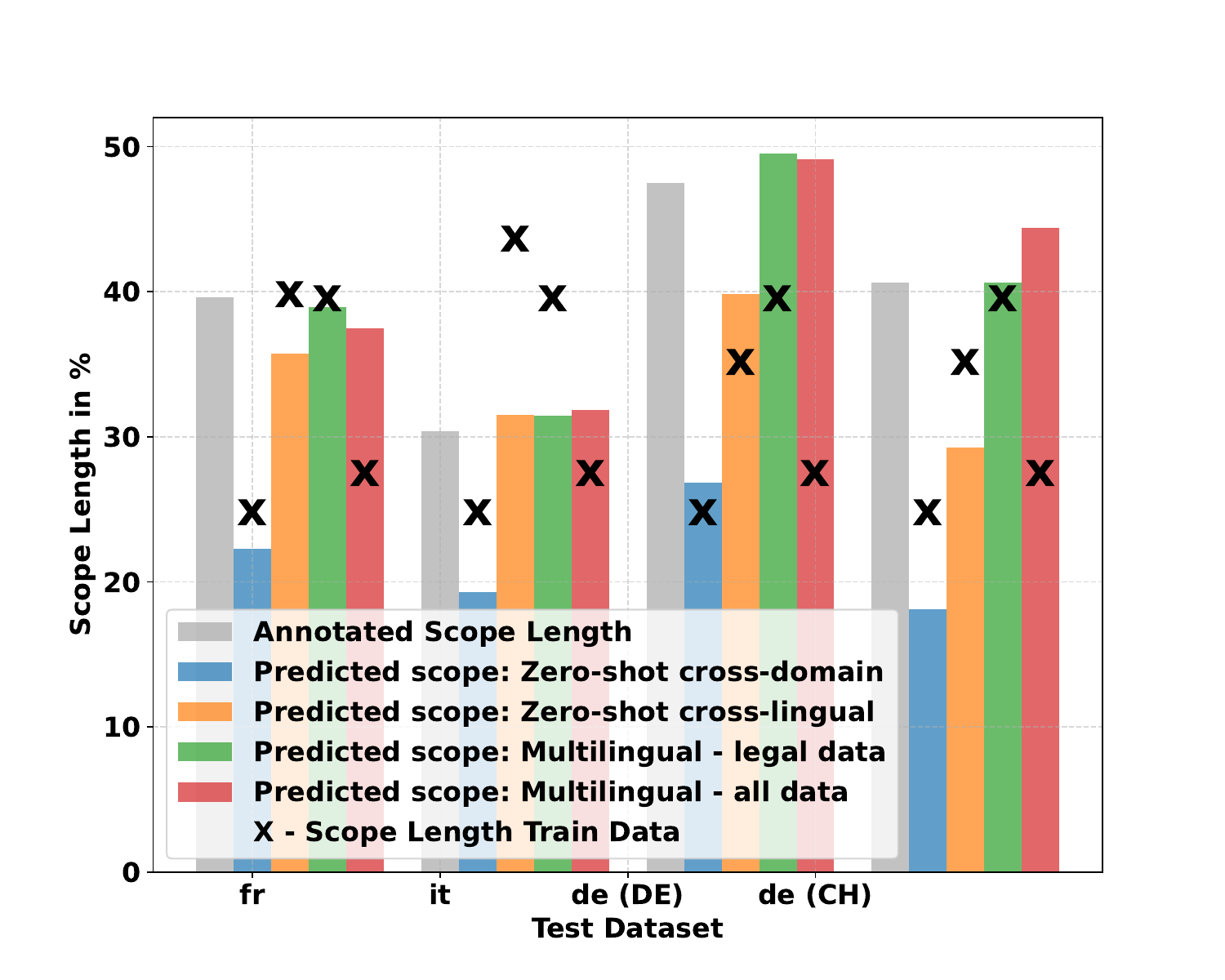}
  \caption{Actual scope length and scope length predicted by Legal-XLM-R\textsubscript{Large} for each experiment. X marks the scope length of the train data. }
    \label{fig:scope_len}
\end{figure}

Analyzing the predicted scope length compared to the actual scope length reveals one main issue with the zero-shot transfer from the external datasets of different domains to our legal datasets. \Cref{fig:scope_len} shows the analysis of the predicted scopes by the Legal-XLM-R\textsubscript{Large} model. In our cross-domain zero-shot experiment, the predicted scope length is significantly shorter than the actual annotated scope length. This is clarified by \Cref{tab:sentence_length}, revealing the external datasets have a shorter annotated scope length (24\%) compared to our legal datasets (38.6\%). Sample predictions confirm that the model often omits the subject from the annotated scope.

\begin{compactitem}
    \item[Annotation]: Es sei festzustellen, dass \ul{der R\"uckerstattungsanspruch} \textbf{nicht} \ul{verj\"ahrt sei}. 
    \item[\textit{EN:}] \textit{It should be noted that \ul{the claim for restitution is} \textbf{not} \ul{forfeited}.}
    \item[Prediction]: Es sei festzustellen , dass der R\"uckerstattungsanspruch \textbf{nicht} \ul{verj\"ahrt sei}.
    \item[\textit{EN:}] \textit{It should be noted that the claim for restitution  \ul{is} \textbf{not} \ul{forfeited}.}
\end{compactitem}

As soon as some legal data is added to our training sets, the predicted scope length as well as the F1-score increases. An inspection of the predictions made by the legal and multilingual models shows that the additional training data helps to predict the subject as part of the scope. One exception where the subject was not annotated in the prediction is for subjects represented by an initial instead of a pronoun or a full name, which is common in legal documents for anonymization reasons. We suspect that in these cases the models were not able to identify the initial as the subject because these kinds of subjects might be more uncommon outside of the legal domain. 
\begin{compactitem}
    \item[Annotation:] \ul{E.\_} \textbf{ne} \ul{disposait d'}\textbf{aucune} \ul{autonomie budg\'etaire};
    \item[\textit{EN:}] \textit{\ul{ E.\_ had} \textbf{no} \ul{budgetary autonomy} }
    \item[Prediction:] E. \_ \textbf{ne} \ul{disposait d'}\textbf{aucune} \ul{autonomie budg\'etaire};
    \item[\textit{EN:}] \textit{E.\_ \ul{had} \textbf{no} \ul{budgetary autonomy} }
\end{compactitem}

\paragraph{Non-continuous scopes}
Another error case is sentences where the scope is not continuous because it is interrupted by an interjection or contrasting statement. These kinds of sentences are more complex than the average sentence and not very common in the training data. A larger amount of training data containing similar sentence structures could improve accuracy.

\begin{compactitem}
    \item[Annotation:] \ul{Eine ordentliche K\"undigung ist w\"ahrend der vereinbarten Laufzeit beiderseits} nur zum Vertragsende und \textbf{nicht} \ul{zu einem fr\"uheren Zeitpunkt zul\"assig}.
    \item[\textit{EN}] \textit{\ul{An ordinary termination during the agreed term is only permissible on both sides} at the end of the contract and \textbf{not} \ul{at an earlier time}}
     \item[Prediction:] \ul{Eine ordentliche K\"undigung ist} w\"ahrend der vereinbarten Laufzeit beiderseits nur zum Vertragsende und \textbf{nicht} \ul{zu einem fr\"uheren Zeitpunkt zul\"assig}. 
      \item[\textit{EN}] \textit{\ul{An ordinary termination} during the agreed term \ul{is only} permissible on both sides at the end of the contract and \textbf{not} \ul{at an earlier time}}
\end{compactitem}

\section{Conclusions and Future Work}
\subsection{Conclusion}
We released new legal datasets in German, French and Italian, annotated for negation cues and scopes and showed that the legal domain does pose a challenge for models in negation scope resolution. Cross-domain zero-shot experiments showed that models without legal training data do not perform as well on multilingual legal datasets as they do on other domains. The task is also too complex for ChatGPT, which was not able to reach F1-scores above 37\%. Using our new datasets we fine-tuned different models on the legal domain, significantly improving the results and showing that even relatively small amounts of training data in a specific domain and language can improve the performance of multilingual \acp{lm} for negation scope resolution.

\subsection{Future Work}
Negation scope resolution models in the legal domain could benefit from more training data to increase the accuracy of predictions of more complex sentence structures such as non-continuous scopes. More diverse data from different legal fields could further improve the performance of negation scope models in the legal domain.

With our new datasets we were able to show that existing systems performing well on datasets across different domains are not necessarily able to perform as well on legal data. This should motivate future work to focus on this complex domain and evaluate the performance of existing systems in diverse NLP tasks.

\section*{Limitations}
Due to resource constraints, our datasets are relatively small compared to other publicly available corpora. A larger set of legal data accross a diverse set of sources, annotated with negations could further improve the performance of \acp{lm} for negation scope resolution in this field. We also did not investigate the potential of cross-lingual cue detection since this is the more trivial part of negation research and can easily be replaced by a list of negation cues for each language.


\section*{Ethics Statement}
The goal of our work was to improve the performance of negation scope resolution systems in the legal domain. These improved systems could be used to support legal professionals in processing and analysing legal texts. These systems should only be used as an assistance to human experts with considerations to their limitations and possible biases. To the best of our knowledge there is currently no real world application of a negation scope resolution system in the legal domain.

The legal data that we annotated and used to train our models is all publicly available and has all been anonymized. It should therefore not include any sensitive information.

\bibliography{bibliography}

\begin{thebibliography}{34}
\expandafter\ifx\csname natexlab\endcsname\relax\def\natexlab#1{#1}\fi

\bibitem[{Altuna et~al.(2017)Altuna, Minard, and Speranza}]{altunaItalian}
Bego{\~n}a Altuna, Anne-Lyse Minard, and Manuela Speranza. 2017.
\newblock \href {https://doi.org/10.18653/v1/W17-1806} {The scope and focus of
  negation: A complete annotation framework for {I}talian}.
\newblock In \emph{Proceedings of the Workshop Computational Semantics Beyond
  Events and Roles}, pages 34--42, Valencia, Spain. Association for
  Computational Linguistics.

\bibitem[{Brown et~al.(2020)Brown, Mann, Ryder, Subbiah, Kaplan, Dhariwal,
  Neelakantan, Shyam, Sastry, Askell, Agarwal, Herbert-Voss, Krueger, Henighan,
  Child, Ramesh, Ziegler, Wu, Winter, Hesse, Chen, Sigler, Litwin, Gray, Chess,
  Clark, Berner, McCandlish, Radford, Sutskever, and
  Amodei}]{brown_language_2020}
Tom Brown, Benjamin Mann, Nick Ryder, Melanie Subbiah, Jared~D Kaplan, Prafulla
  Dhariwal, Arvind Neelakantan, Pranav Shyam, Girish Sastry, Amanda Askell,
  Sandhini Agarwal, Ariel Herbert-Voss, Gretchen Krueger, Tom Henighan, Rewon
  Child, Aditya Ramesh, Daniel Ziegler, Jeffrey Wu, Clemens Winter, Chris
  Hesse, Mark Chen, Eric Sigler, Mateusz Litwin, Scott Gray, Benjamin Chess,
  Jack Clark, Christopher Berner, Sam McCandlish, Alec Radford, Ilya Sutskever,
  and Dario Amodei. 2020.
\newblock \href
  {https://proceedings.neurips.cc/paper/2020/file/1457c0d6bfcb4967418bfb8ac142f64a-Paper.pdf}
  {Language {Models} are {Few}-{Shot} {Learners}}.
\newblock In \emph{Advances in {Neural} {Information} {Processing} {Systems}},
  volume~33, pages 1877--1901. Curran Associates, Inc.

\bibitem[{Brugger et~al.(2023)Brugger, Stürmer, and
  Niklaus}]{brugger_multilegalsbd_2023}
Tobias Brugger, Matthias Stürmer, and Joel Niklaus. 2023.
\newblock \href {https://doi.org/10.48550/arXiv.2305.01211} {{MultiLegalSBD}:
  {A} {Multilingual} {Legal} {Sentence} {Boundary} {Detection} {Dataset}}.
\newblock ArXiv:2305.01211 [cs].

\bibitem[{Chalkidis et~al.(2021)Chalkidis, Fergadiotis, and
  Androutsopoulos}]{multieurlexLegaldata}
Ilias Chalkidis, Manos Fergadiotis, and Ion Androutsopoulos. 2021.
\newblock \href {https://doi.org/10.18653/v1/2021.emnlp-main.559}
  {{M}ulti{EURLEX} - a multi-lingual and multi-label legal document
  classification dataset for zero-shot cross-lingual transfer}.
\newblock In \emph{Proceedings of the 2021 Conference on Empirical Methods in
  Natural Language Processing}, pages 6974--6996, Online and Punta Cana,
  Dominican Republic. Association for Computational Linguistics.

\bibitem[{Chalkidis et~al.(2022)Chalkidis, Jana, Hartung, Bommarito,
  Androutsopoulos, Katz, and Aletras}]{chalkidis_lexglue_2022}
Ilias Chalkidis, Abhik Jana, Dirk Hartung, Michael Bommarito, Ion
  Androutsopoulos, Daniel Katz, and Nikolaos Aletras. 2022.
\newblock {LexGLUE}: {A} {Benchmark} {Dataset} for {Legal} {Language}
  {Understanding} in {English}.
\newblock In \emph{Proceedings of the 60th {Annual} {Meeting} of the
  {Association} for {Computational} {Linguistics} ({Volume} 1: {Long}
  {Papers})}, pages 4310--4330.

\bibitem[{Chapman et~al.(2001)Chapman, Bridewell, Hanbury, Cooper, and
  Buchanan}]{chapman2001}
Wendy~W. Chapman, Will Bridewell, Paul Hanbury, Gregory~F Cooper, and Bruce~G
  Buchanan. 2001.
\newblock \href {https://doi.org/https://doi.org/10.1006/jbin.2001.1029} {A
  simple algorithm for identifying negated findings and diseases in discharge
  summaries}.
\newblock \emph{Journal of biomedical informatics}, 34(5):301--310.

\bibitem[{Conneau et~al.(2020)Conneau, Khandelwal, Goyal, Chaudhary, Wenzek,
  Guzmán, Grave, Ott, Zettlemoyer, and Stoyanov}]{conneau_unsupervised_2020}
Alexis Conneau, Kartikay Khandelwal, Naman Goyal, Vishrav Chaudhary, Guillaume
  Wenzek, Francisco Guzmán, Edouard Grave, Myle Ott, Luke Zettlemoyer, and
  Veselin Stoyanov. 2020.
\newblock \href {http://arxiv.org/abs/1911.02116} {Unsupervised {Cross}-lingual
  {Representation} {Learning} at {Scale}}.
\newblock \emph{arXiv:1911.02116 [cs]}.
\newblock ArXiv: 1911.02116.

\bibitem[{Cotik et~al.(2016)Cotik, Roller, Xu, Uszkoreit, Budde, and
  Schmidt}]{cotikGermanCorpus}
Viviana Cotik, Roland Roller, Feiyu Xu, Hans Uszkoreit, Klemens Budde, and
  Danilo Schmidt. 2016.
\newblock \href {https://aclanthology.org/W16-5113} {Negation detection in
  clinical reports written in {G}erman}.
\newblock In \emph{Proceedings of the Fifth Workshop on Building and Evaluating
  Resources for Biomedical Text Mining {B}io{T}xt{M}2016)}, pages 115--124,
  Osaka, Japan. The COLING 2016 Organizing Committee.

\bibitem[{Dalloux et~al.(2020)Dalloux, Claveau, Grabar, Oliveira, Moro, Gumiel,
  and Carvalho}]{dallouxFrench}
Clément Dalloux, Vincent Claveau, Natalia Grabar, Lucas Oliveira, Claudia
  Moro, Yohan Gumiel, and Deborah Carvalho. 2020.
\newblock \href {https://doi.org/10.1017/S1351324920000352} {Supervised
  learning for the detection of negation and of its scope in french and
  brazilian portuguese biomedical corpora}.
\newblock \emph{Natural Language Engineering}, 27:1--21.

\bibitem[{Devlin et~al.(2019)Devlin, Chang, Lee, and
  Toutanova}]{devlin2018bert}
Jacob Devlin, Ming-Wei Chang, Kenton Lee, and Kristina Toutanova. 2019.
\newblock \href {http://arxiv.org/abs/1810.04805} {{BERT}: {Pre-training} of
  {Deep} {Bidirectional} {Transformers} for {Language} {Understanding}}.
\newblock \emph{arXiv:1810.04805 [cs]}.
\newblock ArXiv:1810.04805.

\bibitem[{Fancellu et~al.(2018)Fancellu, Lopez, and
  Webber}]{fancellu2018multilingual}
Federico Fancellu, Adam Lopez, and Bonnie Webber. 2018.
\newblock \href {http://arxiv.org/abs/1810.02156} {Neural networks for
  cross-lingual negation scope detection}.

\bibitem[{Fancellu et~al.(2017)Fancellu, Lopez, Webber, and
  He}]{fancellu2017neural}
Federico Fancellu, Adam Lopez, Bonnie Webber, and Hangfeng He. 2017.
\newblock \href {https://aclanthology.org/E17-2010} {Detecting negation scope
  is easy, except when it isn{'}t}.
\newblock In \emph{Proceedings of the 15th Conference of the {E}uropean Chapter
  of the Association for Computational Linguistics: Volume 2, Short Papers},
  pages 58--63, Valencia, Spain. Association for Computational Linguistics.

\bibitem[{Friedrich(2021)}]{friedrich2021complexitylaw}
Roland Friedrich. 2021.
\newblock Complexity and entropy in legal language.
\newblock \emph{Frontiers in Physics}, 9:671882.

\bibitem[{Gilardi et~al.(2023)Gilardi, Alizadeh, and
  Kubli}]{gilardi2023chatgpt}
Fabrizio Gilardi, Meysam Alizadeh, and Maël Kubli. 2023.
\newblock \href {https://doi.org/10.1073/pnas.2305016120} {Chatgpt outperforms
  crowd workers for text-annotation tasks}.
\newblock \emph{Proceedings of the National Academy of Sciences},
  120(30):e2305016120.

\bibitem[{Glaser. et~al.(2021)Glaser., Moser., and Matthes.}]{gerDataset}
Ingo Glaser., Sebastian Moser., and Florian Matthes. 2021.
\newblock \href {https://doi.org/10.5220/0010246308120821} {Sentence boundary
  detection in german legal documents}.
\newblock In \emph{Proceedings of the 13th International Conference on Agents
  and Artificial Intelligence - Volume 2: ICAART}, pages 812--821. INSTICC,
  SciTePress.

\bibitem[{ImaniGooghari et~al.(2023)ImaniGooghari, Lin, Kargaran, Severini,
  Jalili~Sabet, Kassner, Ma, Schmid, Martins, Yvon, and Sch{\"u}tze}]{glot500}
Ayyoob ImaniGooghari, Peiqin Lin, Amir~Hossein Kargaran, Silvia Severini,
  Masoud Jalili~Sabet, Nora Kassner, Chunlan Ma, Helmut Schmid, Andr{\'e}
  Martins, Fran{\c{c}}ois Yvon, and Hinrich Sch{\"u}tze. 2023.
\newblock \href {https://aclanthology.org/2023.acl-long.61} {Glot500: Scaling
  multilingual corpora and language models to 500 languages}.
\newblock In \emph{Proceedings of the 61st Annual Meeting of the Association
  for Computational Linguistics (Volume 1: Long Papers)}, pages 1082--1117,
  Toronto, Canada. Association for Computational Linguistics.

\bibitem[{Jim{\'e}nez-Zafra et~al.(2020)Jim{\'e}nez-Zafra, Morante,
  Mart{\'\i}n-Valdivia, and Ure{\~n}a-L{\'o}pez}]{jimenez2020corpora}
Salud~Mar{\'\i}a Jim{\'e}nez-Zafra, Roser Morante, Mar{\'\i}a~Teresa
  Mart{\'\i}n-Valdivia, and L.~Alfonso Ure{\~n}a-L{\'o}pez. 2020.
\newblock \href {https://doi.org/10.1162/coli_a_00371} {Corpora annotated with
  negation: An overview}.
\newblock \emph{Computational Linguistics}, 46(1):1--52.

\bibitem[{Khandelwal and Sawant(2020)}]{khandelwal_negbert_2020}
Aditya Khandelwal and Suraj Sawant. 2020.
\newblock \href {http://arxiv.org/abs/1911.04211} {{NegBERT}: {A} {Transfer}
  {Learning} {Approach} for {Negation} {Detection} and {Scope} {Resolution}}.
\newblock \emph{arXiv:1911.04211 [cs]}.
\newblock ArXiv: 1911.04211.

\bibitem[{Konstantinova et~al.(2012)Konstantinova, De~Sousa, D{\'\i}az,
  L{\'o}pez, Taboada, and Mitkov}]{konstantinova2012sfu}
Natalia Konstantinova, Sheila~CM De~Sousa, Noa P~Cruz D{\'\i}az, Manuel J~Mana
  L{\'o}pez, Maite Taboada, and Ruslan Mitkov. 2012.
\newblock \href
  {http://www.lrec-conf.org/proceedings/lrec2012/pdf/533_Paper.pdf} {A review
  corpus annotated for negation, speculation and their scope}.
\newblock In \emph{Proceedings of the Eighth International Conference on
  Language Resources and Evaluation (LREC'12)}, pages 3190--3195.

\bibitem[{Mackov{\'a} and Straka(2020)}]{uncasedLearningTransfer}
Kate{\v{r}}ina Mackov{\'a} and Milan Straka. 2020.
\newblock Reading comprehension in czech via machine translation and
  cross-lingual transfer.
\newblock In \emph{Text, Speech, and Dialogue}, pages 171--179, Cham. Springer
  International Publishing.

\bibitem[{Mielke(2016)}]{Mie2016Language}
Sabrina~J. Mielke. 2016.
\newblock \href {https://sjmielke.com/acl-language-diversity.htm} {Language
  diversity in {ACL} 2004 - 2016}.

\bibitem[{Morante and Blanco(2012)}]{morante2012sherlock}
Roser Morante and Eduardo Blanco. 2012.
\newblock \href {https://aclanthology.org/S12-1035} {* sem 2012 shared task:
  Resolving the scope and focus of negation}.
\newblock In \emph{* SEM 2012: The First Joint Conference on Lexical and
  Computational Semantics--Volume 1: Proceedings of the main conference and the
  shared task, and Volume 2: Proceedings of the Sixth International Workshop on
  Semantic Evaluation (SemEval 2012)}, pages 265--274, Montr{\'e}al, Canada.
  Association for Computational Linguistics.

\bibitem[{Morante et~al.(2008)Morante, Liekens, and Daelemans}]{morante2008}
Roser Morante, Anthony Liekens, and Walter Daelemans. 2008.
\newblock \href {https://aclanthology.org/D08-1075} {Learning the scope of
  negation in biomedical texts}.
\newblock In \emph{Proceedings of the 2008 Conference on Empirical Methods in
  Natural Language Processing}, pages 715--724, Honolulu, Hawaii. Association
  for Computational Linguistics.

\bibitem[{Morante et~al.(2011)Morante, Schrauwen, and
  Daelemans}]{morante2011annotation}
Roser Morante, Sarah Schrauwen, and Walter Daelemans. 2011.
\newblock Annotation of negation cues and their scope: Guidelines v1.
\newblock \emph{Computational linguistics and psycholinguistics technical
  report series, CTRS-003}, pages 1--42.

\bibitem[{Niklaus et~al.(2021)Niklaus, Chalkidis, and
  Stürmer}]{niklaus_swiss-judgment-prediction_2021}
Joel Niklaus, Ilias Chalkidis, and Matthias Stürmer. 2021.
\newblock \href {https://aclanthology.org/2021.nllp-1.3}
  {Swiss-{Judgment}-{Prediction}: {A} {Multilingual} {Legal} {Judgment}
  {Prediction} {Benchmark}}.
\newblock In \emph{Proceedings of the {Natural} {Legal} {Language} {Processing}
  {Workshop} 2021}, pages 19--35, Punta Cana, Dominican Republic. Association
  for Computational Linguistics.

\bibitem[{Niklaus et~al.(2023{\natexlab{a}})Niklaus, Matoshi, Rani, Galassi,
  Stürmer, and Chalkidis}]{niklaus_lextreme_2023}
Joel Niklaus, Veton Matoshi, Pooja Rani, Andrea Galassi, Matthias Stürmer, and
  Ilias Chalkidis. 2023{\natexlab{a}}.
\newblock \href {https://doi.org/10.48550/arXiv.2301.13126} {{LEXTREME}: {A}
  {Multi}-{Lingual} and {Multi}-{Task} {Benchmark} for the {Legal} {Domain}}.
\newblock ArXiv:2301.13126 [cs].

\bibitem[{Niklaus et~al.(2023{\natexlab{b}})Niklaus, Matoshi, Stürmer,
  Chalkidis, and Ho}]{niklaus_multilegalpile_2023}
Joel Niklaus, Veton Matoshi, Matthias Stürmer, Ilias Chalkidis, and Daniel~E.
  Ho. 2023{\natexlab{b}}.
\newblock \href {http://arxiv.org/abs/2306.02069} {{MultiLegalPile}: {A}
  {689GB} {Multilingual} {Legal} {Corpus}}.
\newblock ArXiv:2306.02069 [cs].

\bibitem[{Niklaus et~al.(2022)Niklaus, Stürmer, and
  Chalkidis}]{niklausJudgmentCorpus}
Joel Niklaus, Matthias Stürmer, and Ilias Chalkidis. 2022.
\newblock \href {http://arxiv.org/abs/2209.12325} {An empirical study on
  cross-x transfer for legal judgment prediction}.
\newblock ArXiv:2209.12325.

\bibitem[{Rasiah et~al.(2023)Rasiah, Stern, Matoshi, Stürmer, Chalkidis, Ho,
  and Niklaus}]{rasiah_scale_2023}
Vishvaksenan Rasiah, Ronja Stern, Veton Matoshi, Matthias Stürmer, Ilias
  Chalkidis, Daniel~E. Ho, and Joel Niklaus. 2023.
\newblock \href {https://doi.org/10.48550/arXiv.2306.09237} {{SCALE}: {Scaling}
  up the {Complexity} for {Advanced} {Language} {Model} {Evaluation}}.
\newblock ArXiv:2306.09237 [cs].

\bibitem[{Ruhl et~al.(2017)Ruhl, Katz, and Bommarito}]{ruhl2017complexitylaw2}
JB~Ruhl, Daniel~Martin Katz, and Michael~J Bommarito. 2017.
\newblock Harnessing legal complexity.
\newblock \emph{Science}, 355(6332):1377--1378.

\bibitem[{Sanh et~al.(2020)Sanh, Debut, Chaumond, and
  Wolf}]{sanh_distilbert_2020}
Victor Sanh, Lysandre Debut, Julien Chaumond, and Thomas Wolf. 2020.
\newblock \href {http://arxiv.org/abs/1910.01108} {{DistilBERT}, a distilled
  version of {BERT}: smaller, faster, cheaper and lighter}.
\newblock \emph{arXiv:1910.01108 [cs]}.
\newblock ArXiv: 1910.01108.

\bibitem[{Shaitarova et~al.(2020)Shaitarova, Furrer, and
  Rinaldi}]{shaitarova_cross-lingual_2020}
Anastassia Shaitarova, Lenz Furrer, and Fabio Rinaldi. 2020.
\newblock \href {https://doi.org/10.5167/UZH-197355} {Cross-lingual
  transfer-learning approach to negation scope resolution}.
\newblock In \emph{CEUR Workshop proceedings}. CEUR-WS.

\bibitem[{Shaitarova and Rinaldi(2021)}]{shaitarova_negation_2021}
Anastassia Shaitarova and Fabio Rinaldi. 2021.
\newblock \href {https://doi.org/10.18653/v1/2021.naacl-srw.3} {Negation
  typology and general representation models for cross-lingual zero-shot
  negation scope resolution in {Russian}, {French}, and {Spanish}.}
\newblock In \emph{Proceedings of the 2021 {Conference} of the {North}
  {American} {Chapter} of the {Association} for {Computational} {Linguistics}:
  {Student} {Research} {Workshop}}, pages 15--23, Online. Association for
  Computational Linguistics.

\bibitem[{Szarvas et~al.(2008)Szarvas, Vincze, Farkas, and
  Csirik}]{szarvas2008bioscope}
Gy{\"o}rgy Szarvas, Veronika Vincze, Rich{\'a}rd Farkas, and J{\'a}nos Csirik.
  2008.
\newblock \href {https://aclanthology.org/W08-0606} {The bioscope corpus:
  annotation for negation, uncertainty and their scope in biomedical texts}.
\newblock In \emph{Proceedings of the workshop on current trends in biomedical
  natural language processing}, pages 38--45. Association for Computational
  Linguistics.

\end{thebibliography}

\bibliographystyle{styles/acl_natbib}

\appendix
\section{ChatGPT negation scope example}
\label{sec:chatGPT}
\textit{User}: In the following German sentence give me a list of all words that are part of a negation scope. Negation scope describes all words affected by a negation in a sentence. Format the output as json. Sentence: Ich weiss nicht was eine Kartoffel ist.

\textit{ChatGPT}:
Certainly, here's the JSON output for the words that are part of the negation scope in the provided German sentence:

\begin{figure}[h]
    \centering
    \includegraphics[width=\columnwidth]{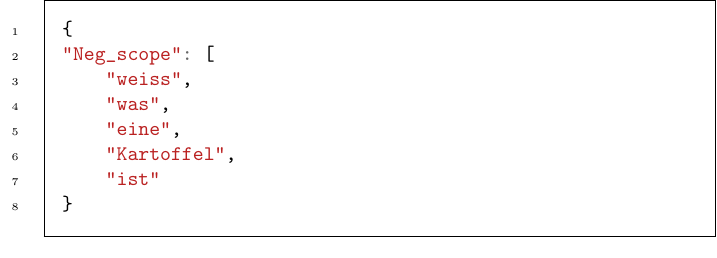}
  \caption{ChatGPT annotation of a simple sentence.} 
\label{json-example-simple}
\end{figure}

\textit{User}:
In the following German sentence give me a list of all words that are part of a negation scope. Negation scope describes all words affected by a negation in a sentence. Format the output as json. Sentence: Die Gew\"ahrung der tariflichen Leistungen, somit auch die Versorgungszusage, die laut Tarifvertrag für Auszubildende nicht gelte, h\"atte \"uberhaupt nur dann Sinn gemacht, wenn die Versorgungszusage entgegen des Geltungsbereichs des Tarifvertrages auch dem Auszubildenden erteilt h\"atte werden sollen.

\textit{ChatGPT}:
Certainly, here's the JSON output for the words that are part of the negation scope in the provided German sentence:
\begin{figure}[h]
    \centering
    \includegraphics[width=\columnwidth]{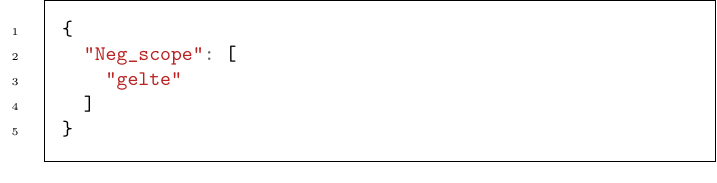}
\caption{ChatGPT annotation of a more complex legal sentence.} 
\label{json-example-legal}
\end{figure}

\onecolumn
\section{Mean results over all models, datasets and experiments}
\label{sec:all_results}

\begin{table}[h]
    \centering
    ~\clap{
 \begin{tabular}{llllll}
 \toprule
                            \diagbox{Model}{Test Dataset}& \Centerstack{Zero-shot\\ cross-domain} & \Centerstack{
                                 Zero-shot\\cross-lingual} 
                            & \Centerstack{
                                 Multilingual -\\ legal data
                            }&    \Centerstack{
                                 Multilingual -\\ all data
                            } & \Centerstack{
                                 Mean F1\\ by model 
                            } \\
\midrule
    DistilmBERT                         &   $61.78_{\pm3.77}$ & $66.46_{\pm13.33}$ & $0.87_{\pm0.05}$ &$88.96_{\pm4.51}$& $72.40_{\pm14.54}$  \\
    mBERT                               &   $63.34_{\pm6.24}$ & $80.33_{\pm7.10}$ &  $0.89_{\pm0.04}$ &$89.19_{\pm4.41}$ & $77.62_{\pm12.33}$  \\
    XLM-R\textsubscript{Base}           &   $66.98_{\pm4.02}$ & $86.71_{\pm3.25}$ & $0.90_{\pm0.03}$ &$91.06_{\pm2.63}$ & $81.59_{\pm11.07}$  \\
    XLM-R\textsubscript{Large}          &   $70.03_{\pm4.98}$ & $85.87_{\pm6.44}$ & $0.90_{\pm0.04}$ &$90.45_{\pm3.00}$ & $82.12_{\pm10.10}$  \\
    \midrule
    Glot500-m                           &   $61.30_{\pm4.85}$ & $84.50_{\pm5.36}$ & $0.89_{\pm0.04}$ &$90.28_{\pm3.84}$ & $78.70_{\pm13.46}$  \\
    \midrule
    Legal-Swiss-R\textsubscript{Base}   &   $66.05_{\pm7.72}$ & $82.79_{\pm7.41}$ & $0.88_{\pm0.05}$ &$89.52_{\pm3.41}$ & $79.45_{\pm11.82}$  \\
    Legal-Swiss-R\textsubscript{Large}  &   $68.89_{\pm10.80}$& $81.83_{\pm7.81}$ & $0.90_{\pm0.03}$ &$90.28_{\pm3.66}$ & $80.33_{\pm11.84}$  \\
    Legal-XLM-R\textsubscript{Base}     &   $66.57_{\pm9.33}$ & $83.44_{\pm7.01}$ & $0.90_{\pm0.04}$ &$89.74_{\pm3.99}$ &$79.92_{\pm12.10}$   \\
    Legal-XLM-R\textsubscript{Large}    &   $71.59_{\pm7.73}$ & $85.25_{\pm4.07}$ & $0.89_{\pm0.04}$ &$90.82_{\pm2.12}$ & $82.55_{\pm9.61}$   \\
    \midrule
    \textbf{Mean F1 by experiment}         &   $66.28_{\pm7.64}$ & $81.91_{\pm9.26}$& $0.89_{\pm0.04}$ & $90.03_{\pm3.57}$ &\\
\bottomrule
\end{tabular}
}
    \caption{Mean Results over all models and experiments}
    \label{tab:all-mean-results}
\end{table}

\section{Multilingual results legal data}
\label{sec:multilingual_legal_results}

\begin{table*}[h]
    \centering
    ~\clap{
    \begin{tabular}{llllll}
    \toprule
                  \diagbox{Model}{Test Dataset} &                fr &                it &                de (DE) &                de (CH) & \Centerstack{Mean F1\\ by Model}\\
    \midrule
    
            DistilmBERT & $86.06_{\pm0.76}$ & $81.82_{\pm0.79}$ & $93.98_{\pm0.82}$ & $87.40_{\pm2.36}$ & $87.32_{\pm4.65}$\\
                  mBERT & $90.16_{\pm1.33}$ & $84.56_{\pm1.63}$ & $94.95_{\pm0.80}$ & $86.81_{\pm2.06}$ & $89.12_{\pm4.25}$\\
             XLM-R-Base & $90.26_{\pm0.96}$ & $88.05_{\pm1.81}$ & $94.12_{\pm0.59}$ & $87.21_{\pm2.66}$ & $89.91_{\pm3.16}$\\
            XLM-R-Large & $90.23_{\pm1.40}$ & $86.93_{\pm0.73}$ & $94.56_{\pm0.85}$ & $86.44_{\pm3.56}$ & $89.54_{\pm3.80}$\\
    \midrule
              Glot500-m & $88.81_{\pm1.47}$ & $85.62_{\pm1.12}$ & $94.23_{\pm1.40}$ & $88.13_{\pm2.60}$ & $89.20_{\pm3.59}$\\
    \midrule
     Legal-Swiss-R-Base & $87.98_{\pm1.46}$ & $89.53_{\pm0.54}$ & $93.15_{\pm0.44}$ & $81.82_{\pm3.88}$ & $88.12_{\pm4.62}$\\
    Legal-Swiss-R-Large & $88.35_{\pm0.88}$ & $88.20_{\pm1.13}$ & $95.30_{\pm0.37}$ & $89.39_{\pm1.37}$ & $90.31_{\pm3.13}$\\
       Legal-XLM-R-Base & $88.89_{\pm1.58}$ & $88.41_{\pm1.84}$ & $95.56_{\pm0.88}$ & $85.27_{\pm3.83}$ & $89.53_{\pm4.39}$\\
      Legal-XLM-R-Large & $88.86_{\pm0.95}$ & $87.98_{\pm0.64}$ & $94.46_{\pm0.69}$ & $85.30_{\pm3.12}$ & $89.15_{\pm3.76}$\\
     \midrule
\textbf{Mean F1 by dataset}&$88.84_{\pm1.70}$&$86.79_{\pm2.55}$ & $94.48_{\pm1.01}$ &$86.42_{\pm3.36}$ &  $89.13_{\pm3.97}$\\
    \bottomrule
    \end{tabular}
    }
    \caption{Results of multilingual experiments using only our legal datasets. }
    \label{tab:mutlilingual-legal-results}
\end{table*}

\end{document}